\documentclass[10pt,twocolumn,letterpaper]{article}

\usepackage{cvpr}
\usepackage{times}
\usepackage{epsfig}
\usepackage{graphicx}
\usepackage{amsmath}
\usepackage{amssymb}
\usepackage{threeparttable}
\usepackage{siunitx}   

\usepackage[breaklinks=true,bookmarks=false]{hyperref}

\cvprfinalcopy 

\def\cvprPaperID{****} 

\ifcvprfinal\fi
\begin{document}

\title{DistanceNet: Estimating Traveled Distance from Monocular Images \\ 
	   using a  Recurrent Convolutional Neural Network}

\author{
	Robin Kreuzig$^1$, \,\,
	Matthias Ochs$^1$, \,\,
	Rudolf Mester$^{2,1}$ \vspace{3pt}\\
	$^1$VSI Lab, Goethe University, Frankfurt am Main, Germany\\
	$^2$Norwegian Open AI Lab, CS Dept. (IDI), NTNU Trondheim, Norway\\
	{\tt\small $^1$\{kreuzig, ochs\}@vsi.cs.uni-frankfurt.de}
	\quad \tt\small $^2$rudolf.mester@ntnu.no
}


\maketitle

\begin{abstract}
Classical monocular vSLAM/VO methods suffer from the scale ambiguity problem. Hybrid approaches solve this problem by adding deep learning methods, for example by using depth maps which are predicted by a CNN. We suggest that it is better to base scale estimation on estimating the traveled distance for a set of subsequent images. In this paper, we propose a novel end-to-end many-to-one traveled distance estimator. By using a deep recurrent convolutional neural network (RCNN), the traveled distance between the first and last image of a set of consecutive frames is estimated by our DistanceNet. Geometric features are learned in the CNN part of our model, which are subsequently used by the RNN to learn dynamics and temporal information. Moreover, we exploit the natural order of distances by using ordinal regression to predict the distance. The evaluation on the KITTI dataset shows that our approach outperforms current state-of-the-art deep learning pose estimators and classical mono vSLAM/VO methods in terms of distance prediction. Thus, our DistanceNet can be used as a component to solve the scale problem and help improve current and future classical mono vSLAM/VO methods.
\end{abstract}
\section{Introduction}

\begin{figure}
  \centering
  \includegraphics[width=0.4\textwidth]{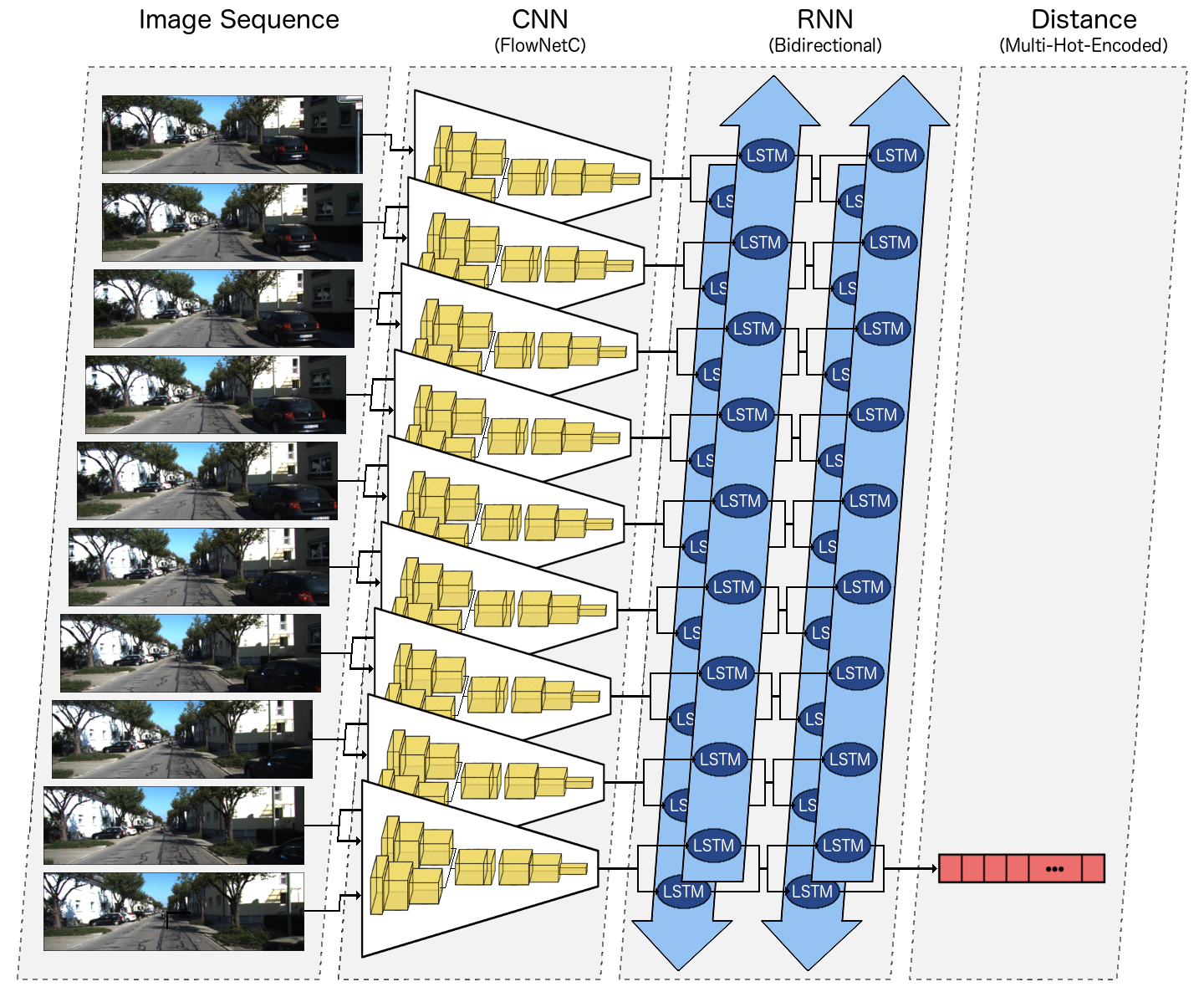}
  \caption{The pipeline of our proposed end-to-end many-to-one DistanceNet. This RCNN architecture consists of a CNN, which learns geometric features and an RNN that infers the traveled distance based on temporal information. The output of our model is a multi-hot-encoded distance class vector.}
  \label{fig:overview}
\end{figure}
   
Autonomous robots and vehicles crucially depend on knowing where they are and how they move in the environment. In the last decade, many variants of SLAM approaches have been proposed to solve exactly this task. SLAM algorithms vary significantly in the types of sensors they use (cameras, Lidar, Radar, ultrasound, ....). Even though most systems deployed to the real world are be based on a fusion of multiple sensors, the demand to push even single sensors to the very limits of their application range is caused by the need to keep systems operational even under partial failure of some sensors. Therefore, there is a current trend in academia as well as in industry to solve the perception tasks only with one monocular camera keeping the same accuracy and robustness.

All monocular visual SLAM (vSLAM) or visual odometry (VO) methods have in common that they may yield state-of-the-art results but cannot observe the absolute scale directly. Thus, they depend on external information (e.g. distance from the camera to the ground plane or measurement of current speed from a speedometer) to resolve this ambiguity. In the present paper, we propose a deep learning approach that determines scale information implicitly during training, without requiring any further external input during operation. The relation between visual motion and 3D motion in the real world is learned during training from pose-annotated visual data (e.g. the KITTI data set).

Our novel deep learning approach estimates the traveled distance of the ego-vehicle exploiting temporal information of consecutive frames. Since the frame rate is a known parameter, the traveled distance is equivalent to the current speed and the scale factor can be directly resolved by simply relating the unscaled speed of a monocular vSLAM or VO method with the absolute scaled traveled distance of our approach. In contrast to that, current hybrid state-of-the-art mono vSLAM approaches apply deep learning techniques to resolve the unknown absolute scale by either deploying CNNs to estimate depth maps from a monocular image or they perform deep learning guided ground plane estimation where a CNN is used to label the ground plane area and external knowledge about the camera height above ground is exploited. In both of these cases, the deep learning method is unable to utilize temporal information and does not directly yield the absolute scale. 

The main contribution of this paper is as follows: We use a \emph{recurrent convolutional neural network} (RCNN) to reliably and robustly determine the traveled distance from a video sequence in an end-to-end manner. For this purpose, we use an optimized RCNN architecture similar to \cite{Wang2017, Wang2018, Jiao2018, Xue2019}, and a novel loss function to train the network. In comparison to other learning based methods, our network outperforms all current state-of-the-art, even classical, mono vSLAM/VO methods in determining the absolute scale. Indeed, one can argue that our approach is simpler than VO networks, which estimate the full 6-DoF pose, but this assumption does not hold, because it makes no difference for estimating the distance if further parameters are also predicted from the given observations. Thus, our proposed DistanceNet can be used to solve the scale problem and even improve classical mono vSLAM/VO method without any additional information.
\section{Related Work}

In this section, we review classical as well as deep learning based vSLAM/VO methods and discuss how they solve the scale ambiguity problem. For the classical approaches, there are basically two possibilities to retrieve the absolute scale: First, it can be resolved indirectly by additional information from an external sensor or it is inferred directly from an appearance based approach, which is mainly based on deep learning techniques. These different types of scale recovery will be studied below.

\subsection{Classical Methods}

Generally, classical methods can be divided into direct and indirect methods. Direct methods, like LSD-SLAM \cite{Engel2014}, DSO \cite{Engel2018}, PMO \cite{Fanani2017} or SVO \cite{Forster2017}, optimize feature correspondences directly in the image by minimizing the photometric error between consecutive frames to retrieve 3D information about the environment and the camera motion simultaneously. 
In contrast to that, indirect or feature-based methods, like PTAM \cite{Klein2007} or ORB-SLAM \cite{Mur-Artal2015}, proceed in two steps: First, some good 2D feature correspondences are found in an image sequence, using one of several proven and tested handcrafted feature descriptor. Then, the 2D coordinates of these correspondences are optimized using a geometric objective function, like the reprojection error, yielding estimates of the geometry and the relative pose. A drawback of all mono vSLAM methods is the unobservable absolute scale and the accumulated scale drift over time. To solve this problem, these approaches depend on external information. For example, this information can be added either in the form of an additional sensor, as a stereo camera \cite{Wang2017a} or a speedometer, or with knowledge about the position of the ground plane relative to the camera without usage of deep learning \cite{Song2014, Zhou2016}.

Recent and state-of-the-art mono vSLAM/VO methods mainly use CNNs to resolve the scale ambiguity problem. The approaches of \cite{Tateno2017, Yin2017, Yang2018} and \cite{Loo2019} have trained a CNN to deploy a scaled depth map from single monocular images. These dense depth maps are used to extend the optimization scheme of the classical frameworks, like for instance a so-called virtual stereo setup \cite{Yang2018} for DSO or initializing depth filters \cite{Loo2019} in SVO directly from the depth map, to eliminate the scale ambiguity. Another possibility, which is pursued by \cite{Fanani2017}, is to first detect the ground plane with a CNN in the image and then to infer the scale with the knowledge of the height of the mounted camera above the detected street level. These hybrid methods represent the current state-of-the-art in terms of accuracy and robustness, which is also confirmed by the rankings in the KITTI odometry benchmark \cite{Geiger2012}.

\subsection{Deep Learning Methods}

\begin{figure*}[t!]
	\includegraphics[width=1.0\textwidth]{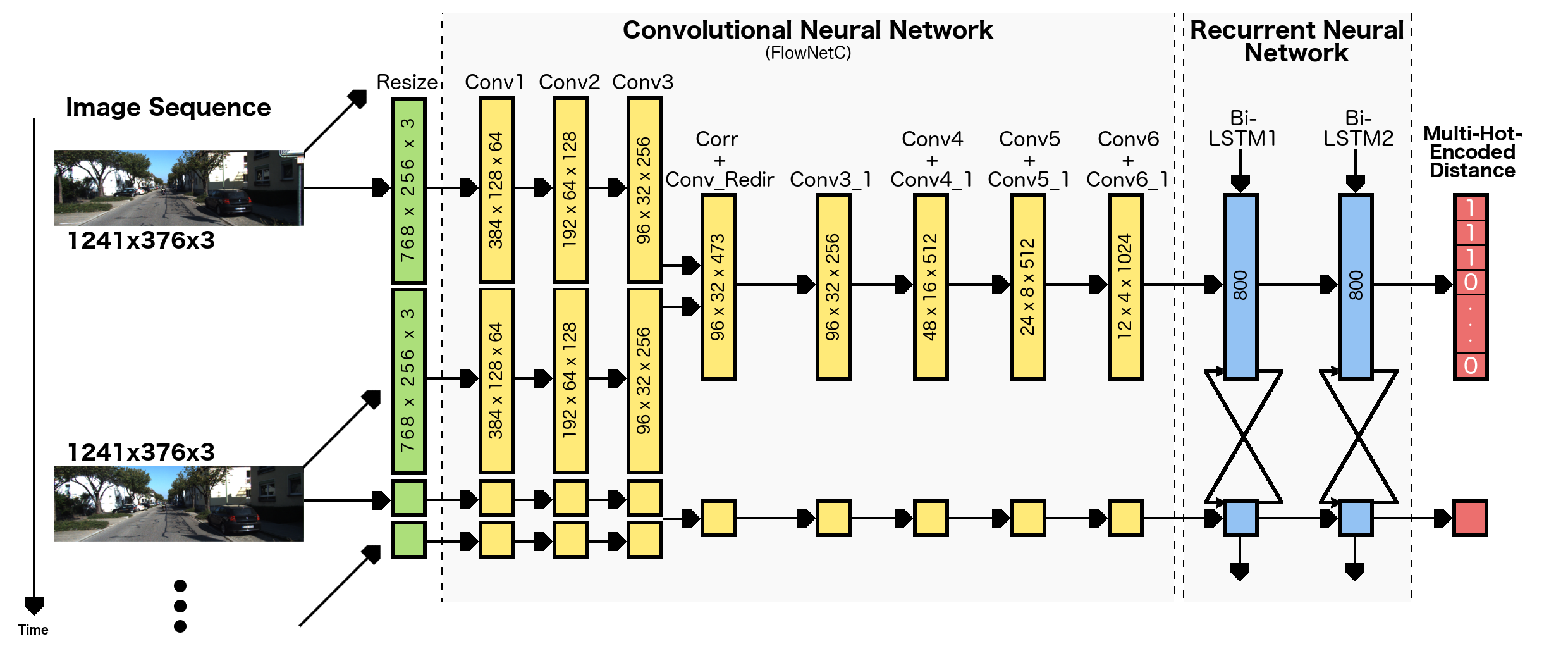}
	\caption{The figure shows the network architecture of our model. After resizing, the images are passed into the CNN. The obtained features are the input for the RNN which returns a multi-hot-encoded distance class.}
	\label{fig:archi}
\end{figure*}

For several years, deep learning approaches have become increasingly successful and achieve astonishing results for solving different computer vision tasks, like image classification, semantic or instance segmentation or even natural language processing. Currently, there are two different end-to-end ways to tackle the vSLAM problem with deep learning.

Unsupervised training of two CNNs, a depth CNN (D-CNN) and a pose CNN (P-CNN) is one possible solution to retrieve the 6-DoF pose with deep learning. In these approaches \cite{Zhou2017, Mahjourian2018, Wang2018a, Zhan2018}, two or three consecutive monocular images or even additional spatial images from a stereo camera jointly serve as the input for both CNNs. Based on this stacked input data, the D-CNN tries to predict a depth map. This depth map and the stacked images are the input of the P-CNN, which estimates a 6-DoF relative pose between these stacked images. This estimated pose and the depth map are used to wrap one (or more, depending on the approach) of the input images that it optimally coincides with the other respectively the reference image. The resulting photometric error between these images and additional penalty terms, like a smoothness term, depending on the approach, form the unsupervised loss function of the CNN which is minimized during training. On the basis of this coincidence of the images, the pose is indirectly predicted by the CNN.
A similar approach was also proposed by \cite{Yin2018} and \cite{Zhao2018}, but they have extended this pipeline with optical flow information. 

A completely different end-to-end approach, which belongs to supervised learning and is quite similar to our proposed method, has been developed in DeepVO \cite{Wang2017} and its successors ESP-VO \cite{Wang2018}, MagicVO \cite{Jiao2018}, DGRNets \cite{Lin2018} and SRNN \cite{Xue2019}. The main idea of these methods is that they estimate the pose based on a temporal sequence of images with a RCNN. In such a network, features are first extracted by a CNN, which is in almost all approaches a variant of the FlowNet \cite{Dosovitskiy2015}, and then passed on as input to a recurrent network consisting of (bidirectional) long short-term memory (LSTM) cells. Within these cells, the network implicitly learns the dynamics and the relations between the input images and finally infers the pose based on this information. A downside to this approach is that no 3D information about the environment can be retrieved. 

Recently, another interesting approach to estimate the 6-DoF pose was published by Almalioglu et al.~\cite{Almalioglu2019}. They predict the pose of consecutive frames from different perspectives with a Generative Adversarial Network (GAN).

Current learning based approaches usually perform worse in comparison to classical vSLAM methods. Additionally, to the best of our knowledge, all current state-of-the-art deep learning techniques that belong to the class of end-to-end methods are only capable of predicting the 6-DoF pose, but cannot reconstruct a temporal consistent map of the environment and do not optimize the pose and map globally like it is done in classical vSLAM methods. Thus, classical approaches are still superior to purly learning based methods and can be further improved, if accurate and reliable scale information is introduced.
\newcommand{\dmax}{d_{\text{max}}}
\newcommand{\dstep}{d_{\text{step}}}

\section{Proposed Model}

In this chapter, we will take a closer look at the architecture of our proposed DistanceNet. The network takes two images as input and returns a vector of multi-hot-encoded distance classes. The model consists of a CNN and a RNN part, as shown in Figure \ref{fig:archi}.
 More details about the two networks are provided in the following sections.

\subsection{Network Architecture}

For calculating the traveled distance between two consecutive images, the network takes these images as input and stacks them together after three convolution layers have been applied to the input images. In a preprocessing step, these images are normalized and resized to a resolution of $768 \times 256$ pixels. 

The CNN is used to extract geometric features of the input, which are semantically meaningful for the RNN to estimate the distance between them. But these features are generally different from the ones of a classification network that is looking for specific image content like a traffic light or pedestrians. Our geometric features are learned over several images and therefore we stack adjacent ones.The obtained features are used as input to the RNN. A recurrent network is capable of learning dynamics and temporal information in video sequences exploiting these features. That means, it is no problem to use several image pairs as input and return the total traveled distance between the first and the last image.

Our network returns a vector of multi-hot-encoded discrete distance classes with the actual class given by the sum of hot-encoded labels.

\subsubsection{Convolutional Neural Network}

The concept of transfer learning can be exploited for a better training convergence and a better generalization to unseen data of the CNN. In order to achieve that, we need to use an already pretrained network that takes two images as input and provides geometric features of the stacked input images. In our approach, we use FlowNetC as feature extractor, which is a variant of FlowNet \cite{Dosovitskiy2015}.

Normally, FlowNetS and FlowNetC are estimating the optical flow between two images. The former takes two stacked images as input while the second processes both separately and concatenates them with a correlation layer afterwards. The architecture including the correlation layer of FlowNetC is shown in Figure \ref{fig:archi}. In this network, the dimension of the input images is shrinked to a resolution of $12 \times 4$  pixels while the feature channel size is increased to $1024$ in the last layer.

\subsubsection{Bidirectional Long Short Term Memory}

Recurrent neural networks have the characteristic to learn through sequences. The RNN gets image pairs over multiple time steps and estimates the distance between the first and last given image. This is possible by passing the previous output as input for the next step whereby information of previous images is stored in memory. This knowledge is used to estimate the traveled distance between consecutive frames.

Unfortunately, this approach suffers from vanishing gradients. Hence, the sequence length must be limited. Using Long Short-Term Memory (LSTM) can be one solution to reduce this problem. A LSTM consist of states, which save information, and gates, which can modify or delete these states. A state with the related gates is called a cell and LSTM layers consist of several of such LSTM cells.

Currently, the LSTM cells can only represent and store information about previous frames. 
But a normal LSTM cell can be extended to a bidirectional LSTM (Bi-LSTM), which does not depend on past data only, but can also process data from current incoming images. For this extension, two standard LSTMs are stacked together. One runs forward as usual and the other backward through the sequence.

In Figure \ref{fig:archi}, the RNN part of our DistanceNet is shown. The model consists of two Bi-LSTM layers with $800$ cells each.

\subsubsection{Output}

\begin{figure}[t!]
  \includegraphics[width=0.5\textwidth]{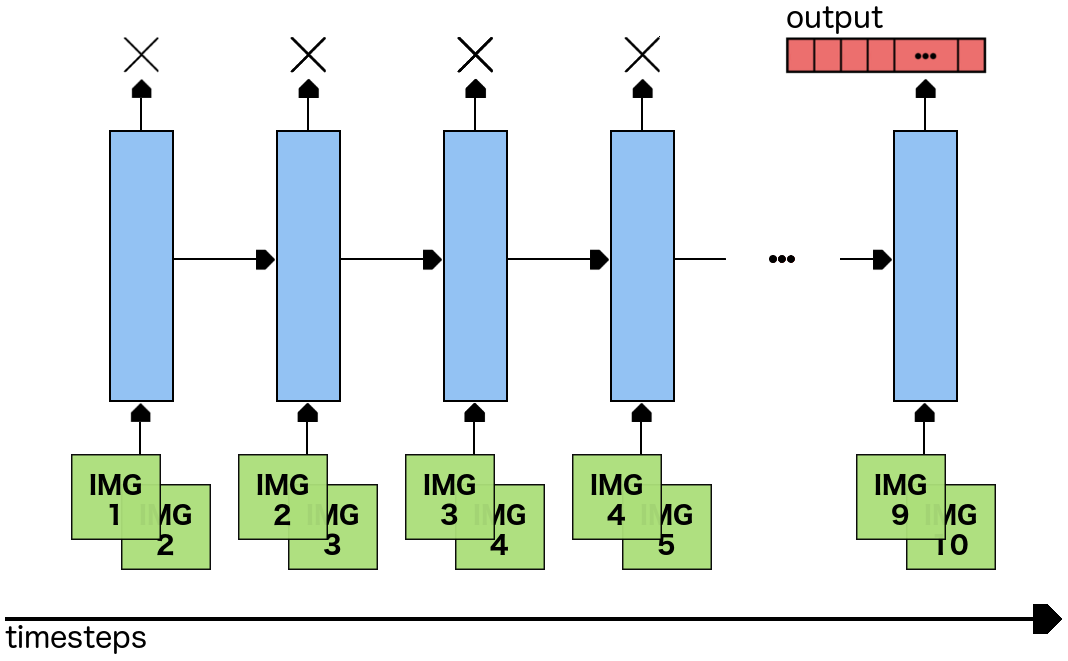}
  \caption{In the many-to-one approach only the output for the last time step matters. The others are thrown away.}
  \label{fig:mto}
\end{figure}

The output of our model is a multi-hot-encoded vector $\vec{v}$ with a length of $K$ discrete distance classes. Unfortunately, in most cases the network does not return integral numbers but probabilities inside the range of $[0, 1]$. Because of that, the output of a single distance class, which is encoded as one component in our output vector, must be rounded to binary labels:
\begin{flalign}
 \vec{v}\ ' &= (\rho(v_{1}), \rho(v_{2}), ..., \rho(v_{K}))^T, \quad \text{with} \\
 \rho(x) &= 
 \begin{cases}
    1, & x \geq t \\
    0, & \text{otherwise}
 \end{cases}
\end{flalign}
Normally, the threshold for rounding is set to $t = 0.5$ but for some models, slightly different thresholds can improve the results.

With just zeros and ones the actual class $c$ can be determined. For that, all the leading ones until the first zero are summed:
\begin{flalign}
 c &= \sum_{k=1}^{K} \eta(\vec{v}\ ', k), \quad \text{with} \\
 \eta(\vec{v}, k) &= 
 \begin{cases}
    1, & k = 1\ \text{and}\ \vec{v}_{k} = 1 \\
    1, & \eta(\vec{v}, k - 1) = 1\ \text{and}\ \vec{v}_{k} = 1 \\
    0, & \text{otherwise}
 \end{cases}
\end{flalign}
The first class that represents a distance of \SI{0}{\metre} is encoded by a vector with only zeros. The opposite with only ones stands for the last class and the maximum distance that can be estimated. Thus, the possible amount of classes determines the granularity of the model. A vector with length of $K$ can encode $K + 1$ classes. With a maximum distance $\dmax$ the distance $\dstep$ between two adjacent classes is calculated by:
\begin{align}
	\dstep = \frac{\dmax}{K}.
\end{align}
The RNN returns these classes for every time step, but it is not necessary to take care of all of them. In our many-to-one approach, which is shown in Figure \ref{fig:mto}, only the last output is required. This means that the network gets image pairs for multiple time steps and given the last pair, it returns the distance between the first and last image.

\subsection{Loss Function and Ordinal Regression}

For a better generalization and training convergence, we divide the traveled distance into classes. These distance classes are predicted by multi-hot-encoded vectors as the output of our network. Moreover, the different classes have a natural order. To keep this information for the network, we can use ordinal regression and transform the distance estimation problem into binary classification subproblems as proposed in \cite{Frank2001}. This means that the distance is no longer estimated on its own. Instead, multiple classifiers are trained and those jointly estimate the result. The training is done by minimizing the mean loss by tuning the hyperparameters $\theta$ of the model with a batch size of $N$, multi-hot-encoded estimations $\vec{e}$ and ground truth multi-hot-encoded classes $\vec{c}$:
\begin{align}
	\theta = argmin_\theta \; \frac{1}{N K} \sum_{n=1}^{N} \sum_{k=1}^{K} L(\vec{e}_{n, k}, \vec{c}_{n, k})
\end{align}
This equation shows that the loss is not only calculated for every element $n$ in the batch (first sum), but rather than that
it is also calculated for every digit $k$ in the element (second sum).
This is where the multiple classifiers are trained. 
As a loss function $L$, we consider \emph{binary cross entropy} (BCE) and
\emph{focal loss} (FL), which was introduced by \cite{Lin2017}.

\subsubsection{Binary Cross Entropy Loss}
In a binary classification problem, two parameters are given. One is the target $t$ that can be $0$ or $1$ and thus encodes the right class that should be predicted. The other one is the probability $p$ in the range of $[0, 1]$, which is the prediction of the network.
For training, we need to measure the error between the target $t$ and the estimated probability $p$. Exactly, this is done with the BCE:
\begin{align}
 \text{BCE}(p, t) = -t\log(p) - (1 - t)\log(1 - p)
\end{align}
The ground truth target $t$ ensures that only one side of the equation is active. A target of $0$ disables the left hand side and a target of $1$ the right. Because of that a good estimation is always approaching the error of  $\log(1) = 0$ and thus means a low error. Vice versa, a bad estimation results in a low value of the logarithm and therefore a high error.

\subsubsection{Focal Loss}
Lin et al. \cite{Lin2017} have published the focal loss an extension to BCE:
\begin{align}
	\text{FL}(p, t) = -t (1-p)^\gamma\log(p) - (1 - t)p^\gamma\log(1 - p)
\end{align}
The left side of the equation measures the error for the positive target $1$. The error of a already well predicted $p$ is reduced by adding $(1-p)^\gamma$ as a multiplier. High values for the prediction $p$ are pushing the multiplier down and thus the complete error. Similarly, the right side with negative target $0$ and low prediction values is doing the same.

This keeps the focus on bad estimations. It is more important to have all values on a sufficient value instead of some perfect ones for the price of a worse.

The effectiveness of the focal loss can be tuned with $\gamma$. As proposed in their paper, we use $\gamma = 2$ and the class balance weights are accordingly mapped into the range of $[0.25, 0.75]$.

\subsubsection{Class Balancing}

Sometimes the amount of the different class examples are unbalanced. This can lead to the network focusing on a few classes only during training, while less common ones are ignored. To avoid that,  we use class balancing weights as follows:
\begin{align}
	L'(p, t) = \alpha_c L(p, t)
\end{align}
The loss function is scaled with the inverted probability of the occurrence $\alpha_c$ of the class $c$ to be estimated. In general, this approach increases the loss of less common classes.

In our model, class weights are applied to full meters. Due to this, the class $c$ must be rounded to get the corresponding $\alpha_c$.

\subsection{Model Parameters}

All important parameters are listed below. In total, 10 normalized and consecutive images with a resolution of $768 \times 256$ pixels are packed into 9 image pairs. These are given to the network as input time step by time step. With the last pair, the network returns the estimated distance between the first and the last of the 10 images.

The CNN architecture of our DistanceNet ist adopted from the FlowNet and pretrained weights are used. Furthermore, we freeze the weights of the first layers (conv\_1, conv\_2, conv\_3, conv\_redir and conv\_3\_1) during training, while the other layers can be updated in the backward pass.

Two Bi-LSTM layers with $800$ cells each form the RNN. To avoid overfitting, a dropout with the rate of $0.3$ is used between and after the layers. In addition, the gradients of the LSTMs are clipped to a value of $1$. Thereby gradient exploding is avoided.

As loss, we use the focal loss with $\gamma = 2$ and the class weights are mapped to the range $[0.25, 0.75]$.
\section{Training and Evaluation}

This section describes the training procedure of our network and the achieved results.
We also compare our results with the pose estimates of other current state-of-the-art vSLAM/VO methods by extracting the estimated distance from them.

\subsection{System Configuration}

We trained our network on the following soft- and hardware:
\begin{itemize}
\item PyTorch 1.0 with Python 2.7.15
\item 3 NVIDIA Titan Xp with 12 GB RAM
\item AMD Ryzen Threadripper 1900X with 64 GB RAM
\end{itemize}

\subsection{KITTI Dataset}

Geiger et al. \cite{Geiger2012} published the KITTI odometry benchmark, which contains of $22$ different driving sequences of a vehicle. 
In order to test the performance of the algorithms in the benchmark, ground truth pose data is only published  for the first $11$ sequences. The sequence $01$ is the only one which is recorded on a highway. We exclude it from our training set, because it comprises only about 1100 images which is too less to generalize to highway scene in contrast to all the other rural and urban sequences, which we use for training our network.

Accumulating the traveled distance of the vehicle in 10 consecutive frames, we set the maximum distance to $ 15 $ meters, which can be split into discrete classes. The KITTI sequences have been recorded with $10$ fps and thus they match exactly with our chosen time step length. That means the output of our network is not only a distance class, but also a speed measurement in meter per second.

\subsection{Training}

Multiple factors influence the results of our network. To show this, we also evaluate the following ablation studies of our model:
\begin{itemize}
	\item DistanceNet-Reg: Regression based training with the mean squared error (MSE) instead of class losses
	\item DistanceNet-BCE: Utilizing BCE loss instead of focal loss
	\item DistanceNet-LSTM: Using standard LSTM cells instead of Bi-LSTMs
	\item DistanceNet-FlowNetS: Replacing the FlowNetC with FlowNetS
\end{itemize}
To train these networks, we first map the pixel values of the RGB images to the range of $ [0, 1] $ and then subtract the pixel means $ [0.411, 0.432, 0.45] $ from them. Furthermore, we use data augmentation to generate more training data for better generalization. Therefore, we flip the images of a input sequence randomly with a probability of $0.5$.

We use Adam as optimizer with a start learning rate of $0.0001$ and beta values of $[0.9, 0.999]$ and weight decay of $0.001$. Every model is trained for up to $200$ epochs. Because the batch size is constrained by the RAM of the GPUs, we use accumulating gradients. The gradients of multiple micro batches are successively calculated and summed before the optimizer is called. With this method, we reach a total batch size of $512$.
   
The output vector length of our model is set to $155$, thus $156$ classes can be encoded in total. This means, the network is estimating in decimeter steps and has a range of \SI{0}{\metre} to \SI{15.5}{\metre}.

\subsection{Evaluation}

\begin{table}
	\begin{center}
	\begin{tabular}{l|cc}
		 Method & $t_{rel}$ & $r_{rel}$ \\
		\hline
		ESP-VO		& 6.16 & 6.66 \\
		DeepVO		& 5.96 & 6.12 \\
		SRNN		& 6.78 & 3.07 \\
		SRNN-se 	& 6.29 & 2.88 \\
		SRNN-point	& 5.47 & 2.53 \\
		SRNN-channel & \textbf{4.97} & \textbf{2.26} \\
		\hline
	\end{tabular}
	\end{center}
	\caption{Averaged translation and rotation error of DeepVO, ESP-VO and different variants of SRNN on the KITTI sequences 03, 04, 05, 06, 07 and 10. The results are taken from the published data of the SRNN paper \cite{Xue2019}.}
	\label{tbl:deepvo}
\end{table}

Three different types of approaches are used to validate our model: 
\begin{itemize}
	\item Classical methods: ORB-SLAM \cite{Mur-Artal2015}, DSO-Mono \cite{Engel2018}, DSO-Stereo \cite{Wang2017a} and PMO \cite{Fanani2017}
	\item CNN methods: GeoNet \cite{Yin2018} and SfMLearner \cite{Zhou2017}
	\item CNN+RNN methods: DeepVO \cite{Wang2017}, ESP-VO \cite{Wang2018} and variants of SRNN \cite{Xue2019}
\end{itemize}
To our knowledge, DeepVO was the first VO system with a RNN structure. The architecture is similar to our network, but uses FlowNetS instead of FlowNetC, LSTMs instead of Bi-LSTMs and poses are estimated instead of discrete distance classes. Furthermore, they are not using ordinal regression. The extended version ESP-VO adds fully connected layers and a $SE(3)$ composition layer at the end of the network. 

Unfortunately, we were not able to get the estimated poses of these networks, but we received the data of the different SRNN variants \cite{Xue2019} published by Xue et al. 
They estimate the rotation and translation parameter of the pose separately and choose suitable features for each of them with guided feature selection. 
For this purpose, SRNN-se uses a SENet \cite{Hu2018} inspired guidance, while SRNN-point is based on point-wise and SRNN-channel channel-wise correlation. 
SRNN does not use guided features at all. To mention briefly, ConvLSTMs \cite{Shi2015} are also used to keep the spatial structure of the features given to a RNN.

The results in table \ref{tbl:deepvo} show that SRNN-point and SRNN-channel performs better than DeepVO and ESP-VO on the averaged translational and rotation error, while SRNN and SRNN-se yield worse results.
With this information, we have upper and lower bounds for the DeepVO and ESP-VO performance and thus it should be possible to compare our results and the ones of them without knowing the estimated poses.

Unfortunately, the different deep learning networks are not trained on the same sequences. Because of that, we evaluate our results in two steps. In the first step, our model and the variants of it are trained on the sequences 00, 02, 08 and 09. By that, we can compare with the different SRNN networks on the common test sequences 03, 04, 05, 06, 07 and 10. The results of all classic methods are added as well.
In the second step, we want to evaluate against GeoNet and SfMLearner. They used as test sequences 09 and 10. Because of that the DistanceNet is trained a second time on the sequences 00, 02, 03, 04, 05, 06, 07 and 08, like these two approaches.

We compare our results on the one hand by using the root mean squared error (RMSE) and on the other hand in terms of the accuracy (Acc) of the predicted right classes of the model. For the latter metric, the estimated and real distances are rounded to full meters and then compared with each other.  A further measure for the consistency of the model is defined by the accuracy with one-meter deviation (AccDev). This means an estimated distance is also classified as correct when it is one meter away from the real distance. A consistent model should have a much higher value on this accuracy than on the accurate one (Acc).

\subsection{Result Analysis}

\begin{table}
	\begin{center}
	\begin{tabular}{l|ccc}
		 Method & RMSE & Acc & AccDev \\
		\hline
		ORB-SLAM-mono & 7.4623 & 0.0221 & 0.0368 \\
		DSO-mono & 7.3854 & 0.0241 & 0.0452 \\
		PMO & 0.7463 & 0.7183 & 0.9633 \\
		DSO-stereo & \textbf{0.0756} & \textbf{0.9387} & \textbf{1.0} \\
		\hline
		\hline
		SRNN & 0.6754 & 0.6121 & 0.9667 \\
		SRNN-se & 0.6526 & 0.5801 & 0.9727 \\
		SRNN-point & 0.5234 & 0.6267 & 0.9822 \\
		SRNN-channel & 0.5033 & 0.6487 & 0.9873 \\
		\hline
		DistanceNet-FlowNetS & 0.5544 & 0.6292 & 0.9752 \\
		DistanceNet-Reg & 0.5315 & 0.6848 & 0.9855 \\
		DistanceNet-LSTM & 0.4167 & 0.6871 & 0.9896 \\
		DistanceNet-BCE & 0.3925 & \textbf{0.7158} & \textbf{0.9930} \\
		DistanceNet & \textbf{0.3901} & 0.6984 & 0.9916 \\
		\hline
	\end{tabular}
	\end{center}
	\caption{Averaged results on KITTI sequences 03, 04, 05, 06, 07 and 10.}
	\label{tbl:results_big}
\end{table}

\begin{table}
	\begin{center}
	\begin{tabular}{l|ccc}
	Method & RMSE & Acc & AccDev \\
	\hline
	GeoNet & 6.2302 & 0.0306 & 0.0544 \\
	SfMLearner & 7.5671 & 0.0216 & 0.0505 \\
	DistanceNet & \textbf{0.4624} & \textbf{0.6669} & \textbf{0.9841} \\
	\hline
	\end{tabular}
	\end{center}
	\caption{Averaged results on KITTI sequences 09 and 10.}
	\label{tbl:results_small}
\end{table}

\begin{figure}
  \centering
  \includegraphics[width=1.0\linewidth]{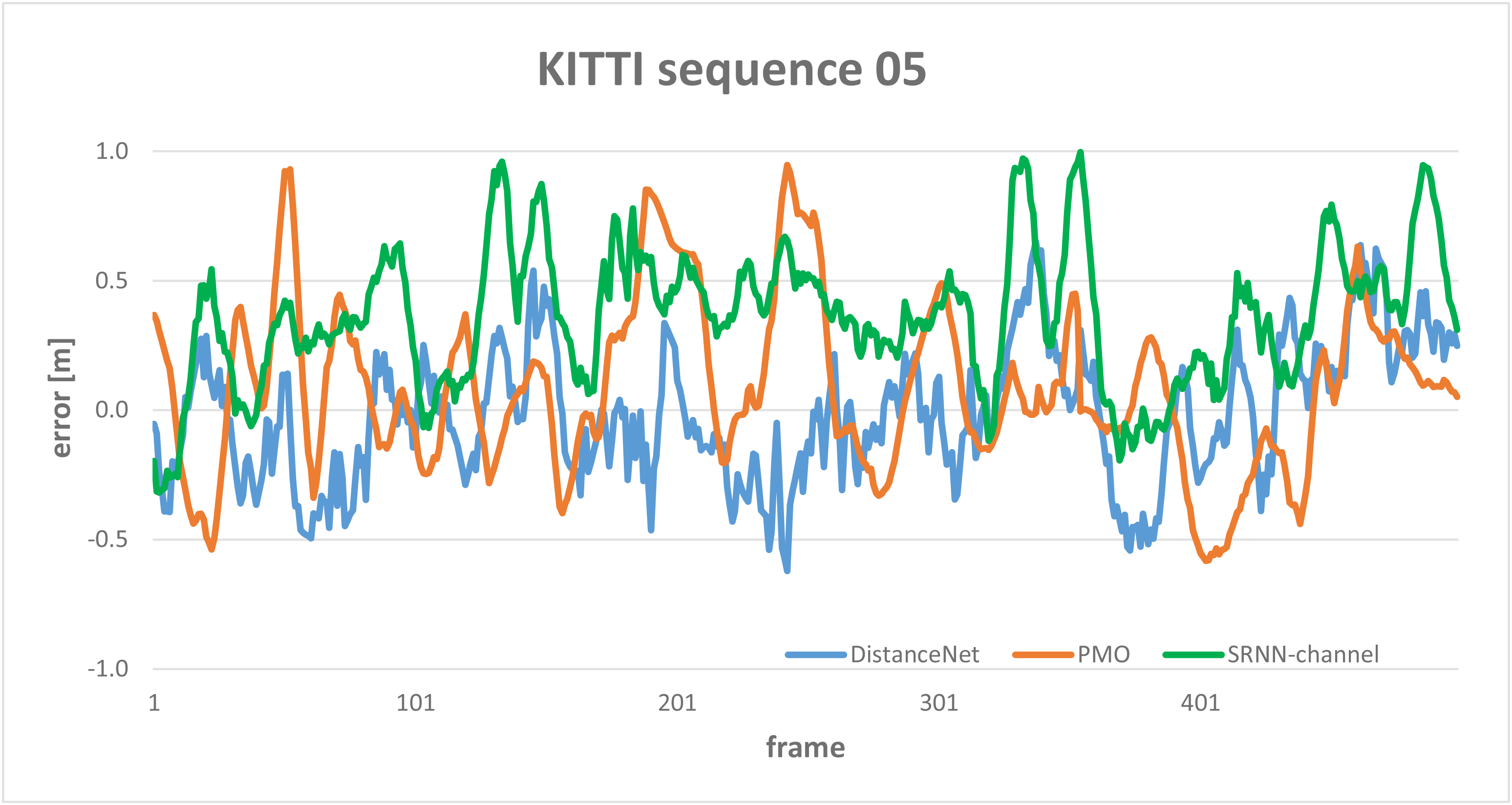} \\
  \vspace{0.2cm}
  \includegraphics[width=1.0\linewidth]{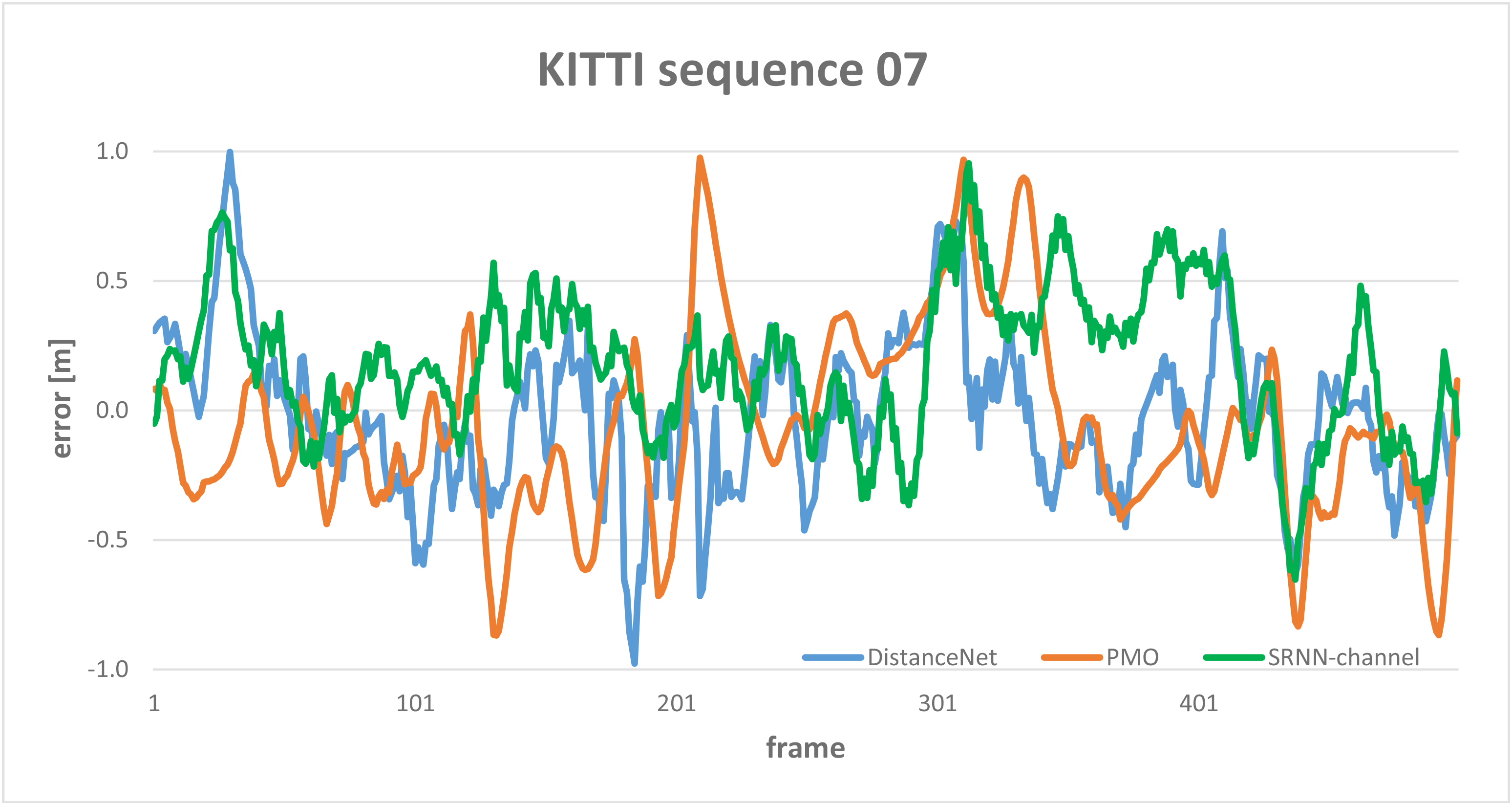}
  \caption{The errors of DistanceNet, SRNN-channel and PMO for the first $500$ frames of the sequences 05 and 07. An error is calculated by subtracting the estimated distance from the actual one. Positive errors exceeded the actual distance and negative ones have not reached it.}
  \label{fig:plots}
\end{figure}

Table \ref{tbl:results_big} is divided into two parts: the upper one for the classic and the lower one for deep learning methods. For both parts, the best results are highlighted. DSO-stereo performs better than anything else but this is not a surprise, because they use a stereo camera setup, where the scale ambiguity does not exist.
The best amongst the classic mono-based methods in our evaluation has been achieved by PMO.

In comparison with our DistanceNet it has a worse result for the RMSE but an unexpectedly high value in accuracy. This is an indicator for an inconsistent method that has some good and some worse results. A view on the accuracy with one-meter deviation confirms this because PMO suddenly falls back.

Since DistanceNet performs better than all SRNN variations on RMSE as well as on accuracy and we declared SRNN-channel as a bound for DeepVO and ESP-VO, we probably perform better than these as well.
The comparison with the networks GeoNet and SfMLearner is shown in Table \ref{tbl:results_small}. Here, the DistanceNet achieved better results as well and thus outperforms all methods without the classical stereo-based DSO-stereo method.

The effect of different network parameters is revealed by our ablation studies of DistanceNet. FlowNetC as well as ordinal regression give a huge performance boost as the comparison between DistanceNet-FlowNetS and DistanceNet-Reg with DistanceNet shows. The usage of Bi-LSTMs is recognizable but not that much and a difference between BCE and focal loss is barely noticeable.

In addition to the results in both tables, Figure \ref{fig:plots} shows the temporal error of DistanceNet, SRNN-channel and PMO over the sequences 05 and 07. A positive error means the actual distance is exceeded. Vice versa, with a negative one it is not reached. All methods suffer from both of these mistakes so that a recognizable pattern cannot be seen. But DistanceNet is always inside the range $[-1, 1]$, actually most of the time it is inside $[-0.5, 0.5]$ and thus better than its counterparts.
\section{Conclusion \& Summary}
In this work, we presented a novel end-to-end deep learning approach for traveled distance prediction on a sequence of consecutive images. Using a RCNN architecture, our approach is able to learn features with a CNN and temporal information with a RNN. Moreover, we exploit the natural order of distances by using ordinal regression. 

The evaluation on the KITTI dataset shows that we outperform current state-of-the-art learning based methods and even classical mono methods for vSLAM/VO. Additionally, we evaluate that better performance is reached by ordinal regression. 

In a next step, we intend to develop a new hybrid mono vSLAM/VO method by incorporating the predicted distance from our DistanceNet to resolve the scale ambiguity problem.

\section*{Acknowledgement}
This project (HA project no. 626/18-49) is financed with funds of LOEWE – Landes-Offensive
zur Entwicklung Wissenschaftlich-ökonomischer Exzellenz, Förderlinie 3: KMU-Verbundvorhaben
(State Offensive for the Development of Scientific and Economic Excellence).

We also gratefully acknowledge the support of NVIDIA Corporation with the donation of the Titan Xp GPU used for this research.

{\small
	\bibliographystyle{ieee}
	\bibliography{wad}
}

\end{document}